\documentclass[preprint,12pt,authoryear]{elsarticle}

\usepackage{amssymb}
\usepackage{amsmath}
\usepackage{graphicx}
\usepackage[lined,boxed,ruled,commentsnumbered]{algorithm2e}
\usepackage{amsfonts}
\usepackage{mathrsfs}
\usepackage{lscape}
\usepackage{float}
\usepackage{subcaption}
\usepackage{caption}
\usepackage{hyperref}
\usepackage{lineno}
\usepackage{lscape}
\usepackage{rotating}

\newcommand{\G}{\mathcal{G}}
\newcommand{\D}{\mathcal{D}}
\newcommand{\X}{\mathbf{X}}
\newcommand{\pa}{\mathit{pa}}
\renewcommand{\S}{\mathcal{S}}
\newcommand{\BLME}{\mathcal{B}_{\mathit{LME}}}
\newcommand{\BCGBN}{\mathcal{B}_{\mathit{CGBN}}}

\journal{Engineering Applications of Artificial Intelligence}

\begin{document}

\begin{frontmatter}

\title{Learning Bayesian Networks with Heterogeneous Agronomic Data Sets via
  Mixed-Effect Models and Hierarchical Clustering}

\author[inst1]{Lorenzo Valleggi}
\author[inst2]{Marco Scutari}
\author[inst3]{Federico Mattia Stefanini}

\affiliation[inst1]{%
  organization={Department of Statistics, Computer Science and Applications,
                University of Florence},
  city={Florence},
  country={Italy}
}
\affiliation[inst2]{%
  organization={Istituto Dalle Molle di Studi sull'Intelligenza Artificiale
               (IDSIA)},
  city={Lugano},
  country={Switzerland}
}
\affiliation[inst3]{%
  organization={Department of Environmental Science and Policy, University of
                Milan},
  city={Milan},
  country={Italy}
}

\begin{abstract}
  Maize, a crucial crop globally cultivated across vast regions, especially in
  sub-Saharan Africa, Asia, and Latin America, occupies 197 million hectares as
  of 2021. Various statistical and machine learning models, including
  mixed-effect models, random coefficients models, random forests, and deep
  learning architectures, have been devised to predict maize yield. These models
  consider factors such as genotype, environment, genotype-environment
  interaction, and field management. However, the existing models often fall
  short of fully exploiting the complex network of causal relationships among
  these factors and the hierarchical structure inherent in agronomic data. This
  study introduces an innovative approach integrating random effects into
  Bayesian networks (BNs), leveraging their capacity to model causal and
  probabilistic relationships through directed acyclic graphs. Rooted in the
  linear mixed-effects models framework and tailored for hierarchical data, this
  novel approach demonstrates enhanced BN learning. Application to a real-world
  agronomic trial produces a model with improved interpretability, unveiling new
  causal connections. Notably, the proposed method significantly reduces the
  error rate in maize yield prediction from 28\% to 17\%. These results advocate
  for the preference of BNs in constructing practical decision support tools for
  hierarchical agronomic data, facilitating causal inference.
\end{abstract}

%%Graphical abstract
\begin{graphicalabstract}
  \includegraphics[width=\linewidth]{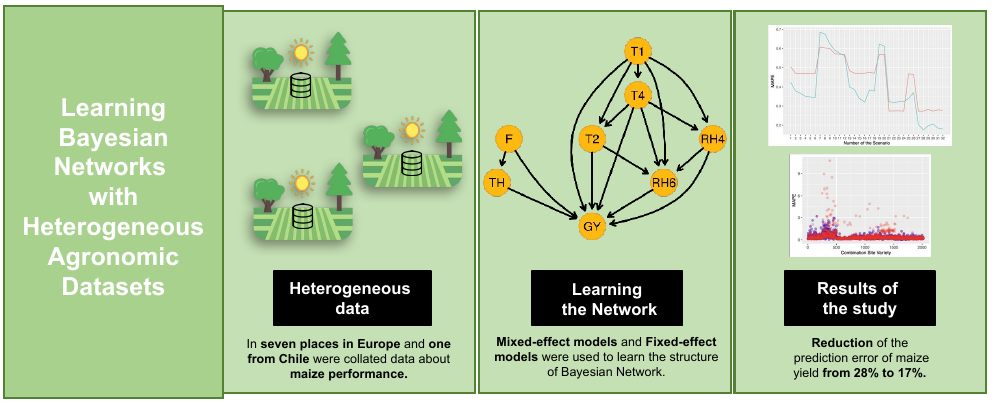}
\end{graphicalabstract}

%%Research highlights
\begin{highlights}
  \item Using mixed-effects models in the learning procedure leads to
    discovering new arcs.
  \item Clustering improves the learning procedure.
  \item Prediction error of the BN maize grain yield decreases approximately
    from 28\% to 17\%.
\end{highlights}

\end{frontmatter}

\section{Introduction}
\label{sec:intro}

The global economy relies on agriculture as a vital source of income and
employment as well as food, ensuring food quality and safety, environmental
preservation, fostering comprehensive rural development, and upholding social
cohesion in rural areas. Given the projected global population growth, which is
expected to reach 9.7 billion by 2050 \citep{pop}, it is estimated that global
agricultural production must increase by 60\% \citep{fao_2012} to meet the
increase in demand. With these premises, improving crop management systems is
essential to match future needs. Maize is one of the most widely cultivated
crops in sub-Saharan Africa, Asia and Latin America, with a total area of 197 M
ha  \citep{fao_state_2021}. It provides almost all the caloric intake in the
Americas (285 kcal/capita/day) and in Africa \citep[374 kcal/capita/day;
][]{faostat}. Predicting the grain yield of this cultivar provides valuable
information about the expected crop output before harvest, enabling more
effective management practices. To achieve accurate predictions, it is essential
to consider the interplay between genotype, environment, and field management.
Widely adopted statistical models for this task include linear mixed-effect and
random coefficient models that use genome-wide association study (GWAS) to study
the causal effects of genotype, environmental variables and their interactions
\citep{zoric_best_2022, ndlovu_genome-wide_2022,tolley_genomic_2023,
rotili_untangling_2020}.

More recently, machine learning models such as random forests
\citep{yang_optimal_2022, leroux_maize_2019}, naive Bayes and SVM
\citep{mupangwa_evaluating_2020} have been applied to maize crop yield
prediction using multi-temporal UAV remote sensing data. Deep learning
architectures such as Long Short-Term Memory \citep[LTSM;
][]{zhang_integrating_2021, krishna_crop_2023} and Convolutional Neural Network
\citep[CNN; ][]{yang_estimation_2021} have also been explored. Despite their
predictive performance, which rests on their ability to encode complex
non-linear relationships, these models are not causal. Outside of randomised
experiments, they are particularly vulnerable to confounding
\citep{pearl_causality}, that is, learning spurious associations as causal
relationships due to unobserved variables acting as common causes of treatment
and outcome and or due to selection bias. They also often disregard the
hierarchical structure that is typical of the data collected in agronomic
studies, which is highly informative.

In general, studies encompassing heterogeneous collections of related data sets
(RDs) in which the relationships between the covariates and the outcome of
interest may differ \citep[say, in slope or variance;][]{gelmandata2007} are
widespread in many fields, from clinical trials to environmental science
\citep{spiegelhalter_bayesian_2004, application_2010}. Hierarchical (multilevel)
models are commonly adopted to pool information across different subsets of the
data while accounting for their specific features \citep{gelman_bayesian_1995}.
However, heterogeneity is not the only challenge in fitting a model on such
data: the variables involved are typically related by a complex network of
causal relationships, making their joint distribution challenging to learn
(especially) from small data sets.

In this work, we chose to learn a Bayesian Network (BN) from RDs focused on the
agronomic performance of maize. BNs can be learned and used as \emph{causal
network models} whose arcs represent cause-effect relationships and which can be
used for causal inference following the work of Judea Pearl
\citep{pearl_causality}. In the case of RDs, learning BNs is also related to
\textit{transfer learning} \citep{pan_survey_2010}, which is not widely
documented in the literature. Transfer learning has mainly focused on
applications involving deep learning, with very few publications involving BNs.
Notably, recent work by \citet{yan_operational_2023} proposed a structure
learning approach based on conditional independence tests for operational
adjustments in a flotation process characterised by a small data set with a
limited sample size. To induce transfer learning, they considered the results of
the independence tests performed on variables $X_i$ and $X_j$ in both the source
and target data sets, which differed in terms of sample size. Other authors have
suggested using order-search algorithms to learn BN structures, introducing
a structural bias term to facilitate the transfer of information between data
sets and achieve more robust networks \citep{oyen_transfer_2015}. BNs and
structural equation models have proven successful in the agronomic sector,
optimising various management practices such as phytosanitary treatments
\citep{lu_disease_2020}, irrigation management strategies
\citep{ilic_irrigation_2022} and soil management \citep{hill_structural_2017},
to minimise environmental impact and mitigate climate change. However, in the
agronomic literature, transfer learning has predominantly focused on crop
disease classification using deep learning techniques like convolutional neural
networks \citep{coulibaly_deep_2019, paymode_transfer_2022}, with little
research involving BNs. A thorough exploration of the literature reveals various
statistical methods for predicting maize grain yield, summarised in
Table~\ref{tab:rev}. We limited ourselves to statistical approaches in keeping
with our focus on modelling frameworks that support causal reasoning.

\begin{table}[p]
  \caption{Statistical methods used for maize grain yield prediction from 2018
    to 2023 in the literature. We reported the data structure, the variables and
    the method used, and whether the method is causal. Abbreviations used:
    greenness index (GI), modified simple ratio (MSR), normalised difference
    vegetation index (NDVI), spectral polygon vegetation index (SPVI), ratio
    vegetation index (RVI), chlorophyll index (CInir), soil-adjusted vegetation
    index (SAVI), triangular vegetation index (TVI), enhanced vegetation index
    (EVI), wide dynamic range vegetation index (WDRVI), temperature, hours of
    sunshine (RAD), Rainfall, Standardised Precipitation Evapotranspiration
    index (SPEI), Peters and Clark Momentary Conditional Independence (PCMCI),
    Linear mixed model (LMM), Linear regression (LM), Hierarchical linear model
    (HLM), Multiple linear regression (MLR), Principal component analysis (PCA),
    Polynomial regression model (PRM), Generalized linear model (GLM), and
    Bayesian network (BN).}
  \label{tab:rev}
  \scriptsize
  \begin{tabular}{p{0.22\linewidth}p{0.25\linewidth}p{0.07\linewidth}lp{0.075\linewidth}p{0.14\linewidth}}
  \hline
  Data structure &
  Variables &
  Method &
  Causal &
  Multi-response &
  Reference \\
  \hline
  period: 2001--2019 \newline locations: 5 \newline trials: 130 &
  FAO maturity groups, Precipitation, Air temperature, Solar radiation, Stress degree days &
  LMM &
  No &
  No &
  \citep{zoric_best_2022} \\
  period: 2011--2015 \newline locations: 3, trials: 13 &
  Soil condition, Genotype data &
  LMM, PCA &
  No &
  No &
  \citep{ndlovu_genome-wide_2022} \\
  period: 2014--2017 \newline locations: 108 \newline plots: 59,416 &
  PAR, Temperature, Humidity, Surface pressure, Wind speed, Precipitation, Soil characteristics, Genotypic data &
  LMM &
  No &
  No &
  \citep{tolley_genomic_2023} \\
  period: 2014--2016 \newline locations: 9 &
  Soil type, Sowing data, Hybrids, Plant density, Row configuration &
  LMM &
  No &
  No &
  \citep{rotili_untangling_2020} \\
  period: 2016--2017 \newline locations: 3, fields: 18 &
  NDVI, NDVIG, NDVIre &
  LM, GLM &
  No &
  No &
  \citep{schwalbert_forecasting_2018} \\
  period: 2000--2016 &
  Temperature, Precipitation, Vapour pressure, Shortwave radiant flux, Soil water content, NDVI &
  MLR &
  No &
  No &
  \citep{kern_statistical_2018} \\
  period: 1981--2016 \newline locations: 12 \newline counties: 1,051 &
  Vapour pressure, Temperature, Precipitation, EVI &
  PRM &
  No &
  No &
  \citep{li_toward_2019} \\
  period: 2014--2018 \newline locations: 23 \newline fields: 94,000 &
  Hybrids, Plant height, Tassel height &
  LMM &
  No &
  No &
  \citep{anderson_ii_prediction_2019} \\
  period: 1986--1987, \newline 2015--2016 &
  Temperature, Precipitation, Heat Magnitude Day, SPEI &
  PCMCI &
  Yes &
  Yes &
  \citep{simanjuntak_impact_2023} \\
  period: 2016--2019 \newline locations: 10 &
  GI, MSR, NDVI, SPVI, RVI, CInir, SAVI, TVI, EVI, WDRVI, Temperature, RAD, Rainfall &
  HLM &
  No &
  No &
  \citep{zhu_regional_2021} \\
  period: 2000--2018 \newline locations: 9, fields: 11 &
  Extreme degree days, Growing degree days, Precipitation &
  Causal forest &
  Yes &
  Yes &
  \citep{kluger_combining_2022} \\
  period: 3 years &
  Temperature, Precipitation &
  BN &
  Yes &
  Yes &
  \citep{ortega_chamorro_urban_2023} \\ \hline
\end{tabular}
\end{table}

\pagebreak

We learned the structure and the parameters of a Conditional Gaussian Bayesian
Network (CGBN) from a real-world agronomic data set with a hierarchical
structure. To account for the high heterogeneity that characterises such data,
we developed a novel approach that integrates random effects into the local
distributions in the BN, building on \citet{scutari2022}. Random effects are the
salient feature of \textit{linear mixed-effects models}
\citep[LME;][]{pinheiro}. LME models are hierarchical models that extend the
classical linear regression model by adding a second set of coefficients called
``random effects'', which are jointly distributed as a multivariate normal. The
other coefficients are called ``fixed effects''. The coefficients associated
with the random effects have mean zero, and they naturally represent the
deviations of the effects of the parents in individual data sets from their
average effects across data sets, represented by the fixed effects.

The hierarchical estimation in BNs learned from RDs was initially introduced by
\citet{azzimonti_hierarchical_2019}, who proposed a novel approach to tackle
this challenge for discrete BNs using a hierarchical multinomial-Dirichlet
model. That approach outperforms a traditional multinomial-Dirichlet model and
is competitive with random forests, but as the number of domains increases, the
estimation becomes more complex, necessitating the use of approximations such as
variational or Markov chain Monte Carlo inference.

The remainder of the paper is structured as follows. In
Section~\ref{sec:methods}, we briefly describe the data set
(Section~\ref{sec:BN}), we introduce the background of BN
(Section~\ref{sec:data}), we introduce the local distributions and the structure
learning approach used to learn the BN (Section~\ref{sec:learn}) and we how we
evaluated its performance (Section~\ref{sec:pred}). In
Section~\ref{sec:results}, we present and evaluate the BN, and in
Section~\ref{sec:discussion}, we discuss its performance before suggesting
possible future research directions.

\section{Materials and Methods}
\label{sec:methods}

\subsection{Background of Bayesian network}
\label{sec:BN}

Bayesian networks \citep[BNs;][]{koller_probabilistic_2009} provide a powerful
tool to learn and model highly structured relationships between variables. A BN
is a  graphical model defined on a set of random variables $\X=\{X_1, \dots,
X_K\}$ and a directed acyclic graph (DAG) $\G$ that describes their
relationships: nodes correspond to random variables, and the absence of arcs
between them implies the conditional independence or the lack of direct causal
effects \citep{pearl_causality}. In particular, a variable $X_i$ is independent
of all other non-parent variables in $\G$ given the set of variables associated
with its parents $\pa(X_i)$. A DAG  $\G$ then induces the following
factorisation:
\begin{equation}
  P(\X \mid \G, \Theta) = \prod_{i=1}^K P(X_i \mid \pa(X_i), \Theta_{X_i}),
\label{eq:parents}
\end{equation}
where $\Theta_{X_i}$ are the parameters of the conditional distribution of $X_i
\mid \pa(X_i)$. In equation (\ref{eq:parents}), the \textit{joint multivariate
distribution} of $\X$ is reduced to a collection of univariate conditional
probability distributions, the \textit{local distributions} of the individual
nodes $X_i$. If all sets $\pa(X_i)$ are small, \eqref{eq:parents} is very
effective in replacing the high-dimensional estimation of $\Theta$ with a
collection of low-dimensional estimation problems for the individual
$\Theta_{X_i}$. Another consequence of \eqref{eq:parents} is the existence of
the \textit{Markov blanket} of each node $X_i$, the set of nodes that makes
$X_i$ conditionally independent from the rest of the BN. It comprises the
parents, the children and the spouses of $X_i$, and includes all the knowledge
needed to do inference on $X_i$, from estimation to hypothesis testing to
prediction.

The process of learning a BN from data can be divided into two steps:
\begin{equation*}
  \underbrace{P(\G,\Theta \mid \D)}_{\text{BN learning}} =
  \underbrace{P(\G \mid \D)}_{\text{structure learning}} \cdot
  \underbrace{P(\Theta \mid \G,\D)}_{\text{parameter learning}}.
\end{equation*}
\textit{Structure learning} aims to find the dependence structure represented by
the DAG given the data $\D$. Several algorithms are described in the literature
for this task. Constraint-based algorithms such as the PC algorithm
\citep{spirtes_causation_2000} use a sequence of independence tests with
increasingly large conditioning sets to find which pairs of variables should be
connected by an arc (or not), and then they identify arc directions based on the
difference in conditional independence patterns between v-structures (of the
form $X_j \rightarrow X_i \leftarrow X_k$, with no arc between $X_j$ and $X_k$)
and other patterns of arcs. Score-based algorithms instead use heuristics
\citep[like hill climbing; ][]{norvig} or exact methods \citep[as in][]{cutting}
that optimise a network score reflecting the goodness of fit of candidate DAGs
to select an optimal one. \textit{Parameter learning} provides an estimate of
$\Theta$ through the parameters in the $\Theta_{X_i}$ conditional to the learned
DAG.

Structure learning algorithms are distribution-agnostic, but the choice of the
conditional independence tests and the network scores depend on the types of
distributions we assume for the $X_i$. The three most common choices are
\textit{discrete BNs}, in which the $X_i$ are multinomial random variables;
\textit{Gaussian BNs} (GBNs), in which the $X_i$ are univariate normal random
variables linked by linear dependence relationships; and \textit{conditional
Gaussian BNs} (CGBNs), in the $X_i$ are either multinomial random variables (if
discrete) or mixtures of normal random variables (if continuous). Common scores
for all these choices are the Bayesian information criterion
\citep[BIC;][]{bic1978} or the marginal likelihood of $\G$ given $\D$
\citep{unification}. As for the conditional independence tests, we refer the
reader to \citet{edwards}, which covers various options for all types of BNs.

Parameter learning uses maximum-likelihood estimates or Bayesian posterior
estimates with non-informative priors for all types of BNs
\citep{koller_probabilistic_2009}. All the conditional independence tests, the
network scores and the parameter estimators in the literature referenced above
can be computed efficiently thanks to \eqref{eq:parents} because they factorise
following the local distributions.

\subsection{The Data Set: Agronomic Performance of Maize}
\label{sec:data}

This study uses the data from \citet{dati_2019}, whose authors are well-known in
plant science research. They conducted a \emph{randomised} genome-wide
association study to assess the genetic variability of plant performance under
different year-to-year and site-to-site climatic conditions. The original
analysis of these data in \citep{millet2016} confirms the quality of this
experimental design in terms of controlling both confounding and various sources
of noise. Overall, $29$ field experiments were arranged in Europe, nine sites,
and in Chile, one site:  each of them was defined by a combination of year, site
and water regime (watered or rain-fed), and $244$ \textit{varieties} of maize
(\textit{Zea mays L.)} were studied. Each experiment was designed as
alpha-lattice design \citep{patterson_new_1976}, with two replicates of the
watered regime and three for the rain-fed regime. The data were collected at
experimental sites in France, Germany, Italy, Hungary, Romania, and Chile
between 2011 and 2013. After filtering out incomplete observations, the study
analysed eight \textit{sites}, each with a different sample size: Gaillac
(France, $n=2437$), Nerac (France, $n=1716$), Karlsruhe (Germany, $n=2626$),
Campagnola (Italy, $n=1260$), Debrecen (Hungary, $n=2181$), Martonvasar
(Hungary, $n=1260$), Craiova (Romania, $n=1055$), and Graneris (Chile, $n=760$).
Many weather variables were measured for each site, such as air temperature,
relative humidity (RH), wind speed and light; they were measured every hour at
2$m$ height. Soil water potential was measured daily at 30, 60, and 90$cm$
depths. For this analysis, we decided to use only air temperature and RH because
they are the more basic variables that can describe plant growth. Since weather
variables were measured for each site instead of for each plot, we decided to
aggregate the weather data in order to have the \textit{average temperature}
($^{\circ}C$), the \textit{diurnal temperature range}, the \textit{average
relative humidity} (\%) and the \textit{diurnal relative humidity range} (\%)
for each site and year for three different periods, which correspond to the main
phenological stages of maize: seeding, germination, and emergence of the seeds,
the vegetative phase, where leaves emerge (May to June), the flower development,
pollen shedding, grain development (July to August), maturation of the grain,
and harvest (September to October). Furthermore, random noise with a mean of 0
and a standard deviation of 0.1 was added to each weather observation to
simulate the sensor's measurement error and avoid blocks of identical
measurements in each individual site.

At the end of the experiment, the phenological variables listed below were
measured for each plot at each site:
\begin{itemize}
  \item The \textit{grain yield} adjusted at 15\% grain moisture, in ton per
    hectare ($t / ha$).
  \item The \textit{grain weight} of individual grains ($mg$).
  \item The \textit{anthesis}, male flowering (pollen shed), in thermal time
    cumulated since emergence (d20$^{\circ}$C).
  \item The \textit{sinking}, female flowering (silking emergence), in thermal
    time cumulated since emergence (d20$^{\circ}$C).
  \item The \textit{plant height} from ground level to the base of the flag leaf
    (highest) leaf ($cm$).
  \item The \textit{tassel height}, plant height including tassel, from ground
    level to the highest point of the tassel ($cm$).
  \item  The \textit{ear height}, ear insertion height, from ground level to the
    ligule of the highest ear leaf ($cm$).
\end{itemize}

\subsection{Learning Algorithm}
\label{sec:learn}

We learned the structure of the BN, denoted $\BLME$, following the steps in
Algorithm~\ref{alg:learning}.

\begin{algorithm}[ht]
  \caption{Structure learning $\BLME$.\label{alg:learning}}
  \KwData{data set $\D$, $blacklist$, and a $whitelist$}
  \KwResult{ The DAG $\G_{max}$ that maximises BIC($\G_{max}$ ,$\D$).}
  \vspace{0.5\baselineskip}
  \begin{enumerate} \setlength{\itemindent}{-0.5em}
    \item Run a linear regression on grain yield and extract the residuals
      $\epsilon_i$.
    \item For each Site $\times$ Variety combination, compute the mean and the
      standard deviation of $\epsilon_i$.
    \item Perform hierarchical clustering on the means and standard deviations
      of the residuals from each site-variety combination.
    \item Add a new variable with the cluster labels to $\D$.
    \item Compute the score of $\G$, \mbox{$\S_{\G} = \mathrm{BIC}(\G , \D)$}
      and set \mbox{$\S_{max} = \S_{\G}$} and \mbox{$\G_{max} = \G$}.
    \item \textit{Hill-climbing}: repeat as long as $\S_{max}$ increase:
    \begin{enumerate} \setlength{\itemindent}{-0.5em} \item Add, delete or
        reverse all possible arc in $\G_{max}$ resulting in a DAG.
      \begin{enumerate} \setlength{\itemindent}{-0.5em} \item compute BIC of the
          modified DAG $\mathcal{G^*}$, \mbox{$\S_{G^*}=\mathrm{BIC}(\G^*,
        \D$)}; \item if \mbox{$\S_{G^*} > \S_{max}$} and \mbox{$\S_{G^*} > \S$}
          set  \mbox{$\G = \G_{^*}$} and \mbox{$\S_{\G} = \S_{\G_{^*}}$}.
      \end{enumerate} \item Update $\S_{max}$ with the new value of
    $\S_{\G_{^*}}$. \end{enumerate} \item Return the DAG $\G$. \end{enumerate}
  \end{algorithm}

For the hill-climbing algorithm, we used the implementation in the $bnlearn$ R
package \citep{bnlearn} and the BIC score. We provided a list of arcs to be
excluded (\textit{blacklist}) or included (\textit{whitelist}) by hill-climbing
to avoid evaluating unrealistic relationships (such as the Average temperature
of July--Aug $\to$ Average temperature of May--June).

Firstly, we regressed the grain yield against all the available variables for
all combinations of site and variety. We used the residuals' mean and variance
from the regression for each combination of site and variety to cluster them
using the agglomerative Ward clustering algorithm \citep{wards_2014} from the
$stats$ R package. The resulting discrete variable was added to the data used
to identify the RDs.

Following the approach and the notation described in \citet{scutari2022}, we
assumed that each RD is generated by a GBN and that all GBNs share a common
underlying network structure but different parameter values. To ensure
the partial pooling of information between RDs, the clusters are made a common
parent for all phenological variables and incorporated into the local
distributions as a random effect. Therefore, we modelled the local distributions
for those variables as a linear mixed-effect model using the $lme4$ R package
\citep{bates_fitting_2014}:
\begin{align}\label{eq:1}
  X_{i, j} = (\mu_{i, j} + b_{i, j, 0}) +
    \boldsymbol{\Pi}_{X_i}(\beta_i + b_{i, j}) + \epsilon_{i, j}, \\ \nonumber
  \begin{pmatrix} b_{i, j, 0} \\ b_{i, j} \end{pmatrix}
  \sim N(\boldsymbol{0},\boldsymbol{\widetilde{\Sigma}}_i), \\ \nonumber
  (\epsilon_{i, 1}, \ldots, \epsilon_{i, j}, \ldots)^T \sim
    N(\boldsymbol{0}, \sigma^2_i \boldsymbol{I}_{nj})
\end{align}
where bold letters indicate matrices. The only exception was grain yield because
it also required a model for variances, which have been implemented using $nlme$
R package \citep{nmle2017} as follows:
\begin{align}\label{eq:2}
  X_{i, j} = (\mu_{i,j}+b_{i, j, 0}) +
    \boldsymbol{\Pi}_{X_i}\beta_i + \epsilon_{i, j}, \\ \nonumber
  b_{i, j, 0} \sim N (0, \sigma^2_{b, i}), \\ \nonumber
  N(0, (\sigma^2_{i, 1}, \sigma^2_{i, 2}, \ldots, \sigma^2_{i, j}, \ldots)
    \boldsymbol{I}_{nj}), \\ \nonumber
 (\epsilon_{i, 1}, \ldots, \epsilon_{i, j}, \ldots)^T \sim
  N(\boldsymbol{0}, (\sigma^2_{i, 1}, \ldots, \sigma^2_{i, j}, \ldots)
    \boldsymbol{I}_{nj}), \\ \nonumber
    \sigma_{i, j}^2(\nu) = \left | \nu \right |^{2\theta_j}.
\end{align}
In both \eqref{eq:1} and \eqref{eq:2}, the notation is as follows:
\begin{itemize}
  \item $j = 1, \ldots, J$ are the clusters identifying the RDs;
  \item $\boldsymbol{\Pi}_{X_i}$ is the design matrix associated to the parents
    of $X_i$;
  \item $b_{i, j, 0}$ is the random intercept;
  \item $b_{i, j}$ is the random slope parameter for the $j$th cluster;
  \item $\boldsymbol{\widetilde{\Sigma}}_i$ is the $n_j \times n_j$ block of
    $\boldsymbol{\Sigma}_i$ associated with the $j$th cluster;
  \item $\sigma^2_{i, j}\boldsymbol{I}_{nj}$ is the $n_j \times n_j$ matrix
    arising from the assumption that residuals are homoscedastic in \eqref{eq:1};
  \item $\mu_{i, j}$ is the intercept;
  \item and $\beta_i$ are the fixed effects.
\end{itemize}
In \eqref{eq:2}, we assumed the variance of residuals to be heteroscedastic and
following a power function, where $\nu$ is the variance covariate and $\theta_j$
is the variance function coefficient that changes for every level of the common
discrete parent.

We modelled the weather variables using only fixed effects for simplicity:
\begin{align}
  &X_{i} = \mu_{i} + \boldsymbol{\Pi}_{X_i}\beta_i + \epsilon_i,&
  &\epsilon_i\sim N(0,\sigma^2_i~ \boldsymbol{I}_n).
\label{eq:3}
\end{align}
We prevented the clusters from being their parent with the blacklist because the
resulting arcs are not of interest from an agronomic perspective.

From these assumptions, the BN $\BLME$ we learned from the data has a global
distribution that is a mixture of multivariate normal distributions like a CGBN.

\subsection{Predictive and Imputation Accuracy}
\label{sec:pred}

The most important variable in this analysis was grain yield because it is one
of the key quantities used to evaluate an agronomic season. To assess the
predictive ability of $\BLME$, we evaluated the Mean Absolute Percentage Error
(MAPE) of:
\begin{itemize}
  \item the \textit{predictive accuracy} of grain yield predictions in the
    scenarios listed in Table~\ref{tab:scenarios}, which are meant to study the
    potential effect of measuring a reduced set of variables in future years;
  \item the \textit{imputation accuracy} for the grain yield in each
    site-variety combination, which we removed in turn and imputed from the
    rest.
\end{itemize}
As a term of comparison, we used a CGBN learned from the data ($\BCGBN$) and
compared its performance with that of $\BLME$. We implemented both prediction
and imputation using likelihood weighting
\citep{koller_probabilistic_2009,reasoning_2009}.

To validate the learning strategy in Algorithm~\ref{alg:learning}, we performed
50 replications of hold-out cross-validation where 20\% of the site-variety
combinations were sampled and set aside to be used as a test set. The remaining
80\% was used as a training set to learn $\BLME$ and $\BCGBN$. We computed the
predictions for each phenological node $X_i$ (except for grain yield) from its
Markov blanket and used these predictions to predict the grain yield in turn. We
used the kernel densities of the predicted values and the resulting credible
intervals with coverage $0.80$ to assess the variability in prediction.

We are aware that predictive accuracy is not an adequate performance measure
for a causal model, which is why we also validated it using expert knowledge
from the literature in Section~\ref{sec:discussion}. However, it allows for
comparisons with other machine learning models that cannot be assessed
causally, and it can be used to evaluate specific loss functions for $\BLME$ in
different application settings. A visual summary of the above data analysis is
reported as a mechanistic diagram as Supplementary Material.

\section{Results}
\label{sec:results}

\begin{sidewaysfigure}[htbp]
  \centering
  \includegraphics[width=0.9\linewidth]{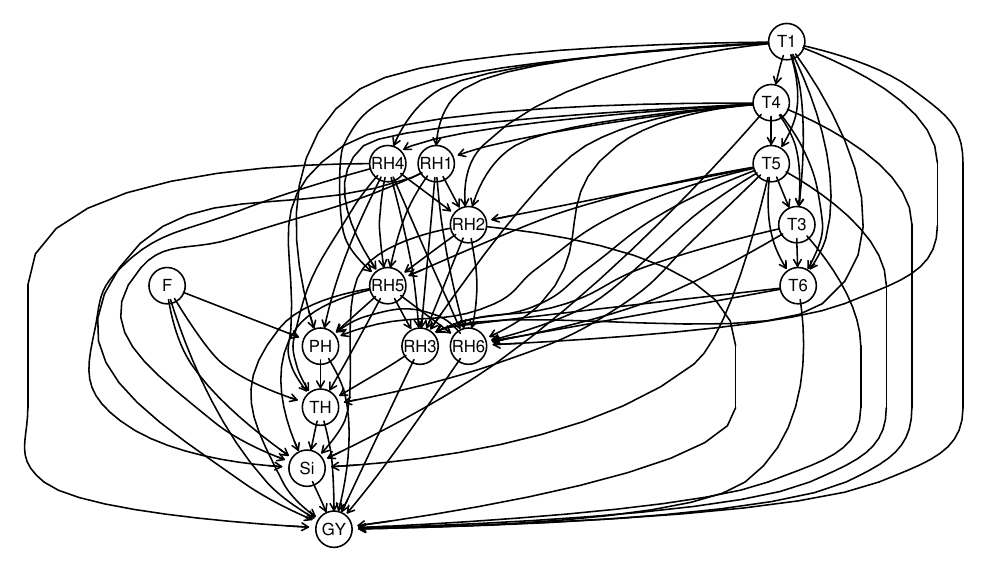}

  \caption{Structure of the BN $\BLME$ learned with Algorithm~\ref{alg:learning}.
  The nodes represent: the average temperature May--June (T1),
  the average temperature Sept--Oct (T3),
  the diurnal temperature range May--June (T4),
  the diurnal temperature range July--Aug (T5),
  the diurnal temperature range Sept--Oct (T6),
  the average RH May--June (RH1),
  the average RH July--Aug (RH2),
  the average RH Sept--Oct (RH3),
  the diurnal RH range May--June (RH4),
  the diurnal RH range July--Aug (RH5),
  the diurnal RH range Sept--Oct (RH6),
  Silking (Si),
  TH (Tassel height),
  PH (Plant height),
  EH (Ear height) and
  F (Clusters).}
  \label{fig:blme}
\end{sidewaysfigure}

\begin{sidewaysfigure}[htbp]
  \centering
  \includegraphics[width=0.9\linewidth]{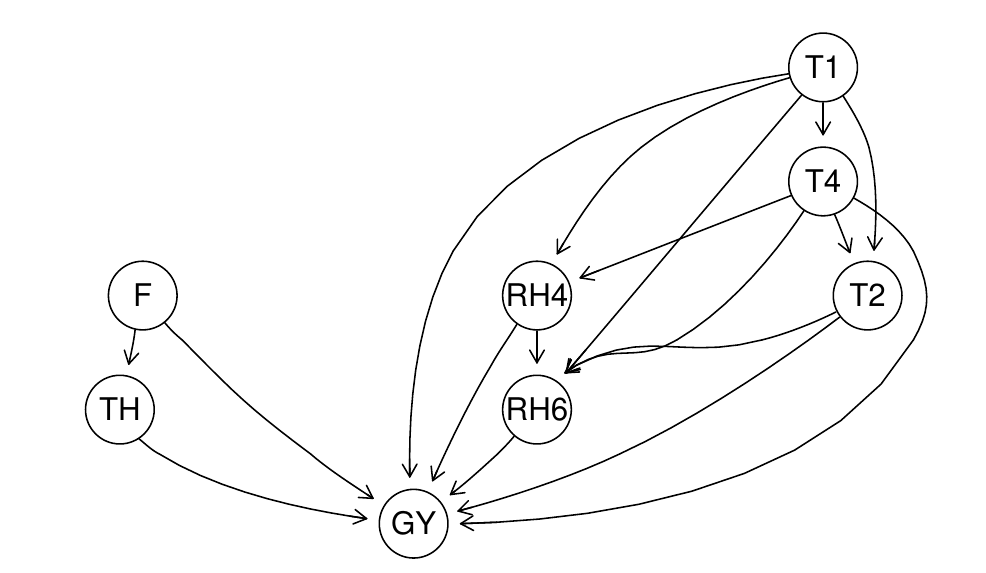}

  \caption{Structure of the BN $\BCGBN$.
    The nodes represent: the average temperature May--June (T1),
    the average temperature July--Aug (T2),
    the diurnal temperature range May--June (T4),
    the diurnal RH range May--June (RH4),
    the diurnal RH range Sept--Oct (RH6),
    TH (Tassel height) and
    F (Clusters).}
  \label{fig:bcgn}
\end{sidewaysfigure}

The complete BNs $\BLME$ and $\BCGBN$ learned from the data are shown in the
Supplemental Material. Here, we show only the subgraph around the variable grain
yield for each BN in Figure~\ref{fig:blme} and~\ref{fig:bcgn}. Following
Section~\ref{sec:learn}, we identified 60 site-variety clusters (with only 5
containing fewer than 100 observations) and used them as a discrete variable set
to be the parent of the phenological nodes.

The structure of $\BLME$ is more complex than that of $\BCGBN$: $\BLME$ has 118
arcs compared to the 92 of $\BCGBN$, and the average Markov blanket size
reflects that  (17 for $\BLME$, 12 for $\BCGBN$). Notably, we discovered more
relationships for the phenological nodes, particularly for the grain yield
variable (Table~\ref{tab:arc}), which had eight more parents than in $\BCGBN$.

\begin{figure}[H]
 \hspace{-1cm}
 \includegraphics[width=14.5cm,height=9.5cm]{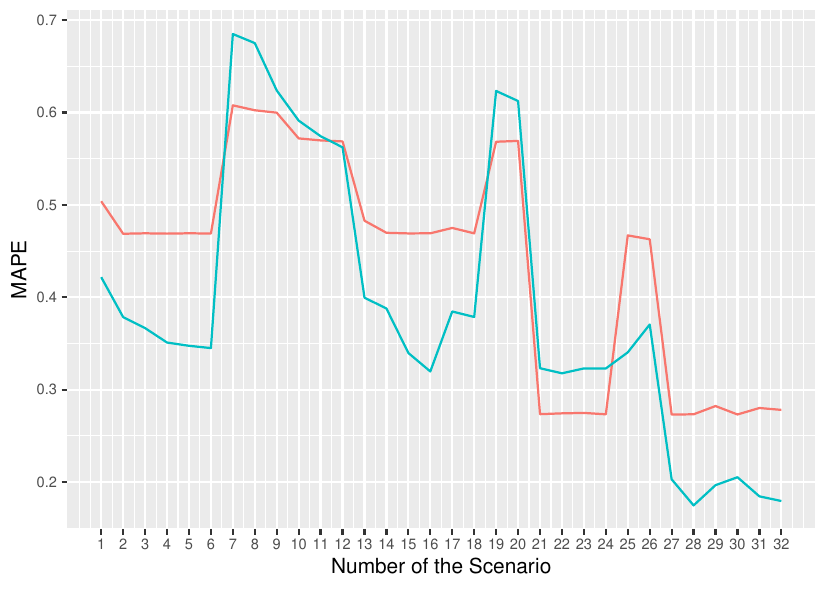}
 \caption{Prediction accuracy of the learned BNs, $\BLME$ (blue line) and
   $\BCGBN$ (orange line), in terms of grain yield Mean Absolute Percentage
   Error (MAPE) of each scenario of evidence propagation (definitions of the
   scenarios are reported in Table~\ref{tab:scenarios}). Lower values are
   better.}
  \label{fig:accuracy}
\end{figure}

The predictive accuracy for each of the scenarios reported in
Table~\ref{tab:scenarios} is shown in Figure~\ref{fig:accuracy} for both $\BLME$
and $\BCGBN$. Overall, $\BLME$ outperformed $\BCGBN$ in terms of MAPE. The
exception was in a few cases, specifically scenarios 7 to 11, 19, 20 and from 21
to 24, where $\BCGBN$ demonstrated a lower MAPE than $\BLME$, albeit with a
difference in MAPE of only 0.06. In contrast, when $\BLME$ outperformed
$\BCGBN$, the difference in MAPE was 0.14. This trend was particularly evident
in scenarios 27 to 32, where an increasing usage of weather/phenological
variables was provided. As expected, the scenarios with the lowest MAPE utilised
the Markov Blanket (scenario 31) and the parents of grain yield (scenario 32).

\begin{figure}[H]
\centering
  \includegraphics[width=\textwidth]{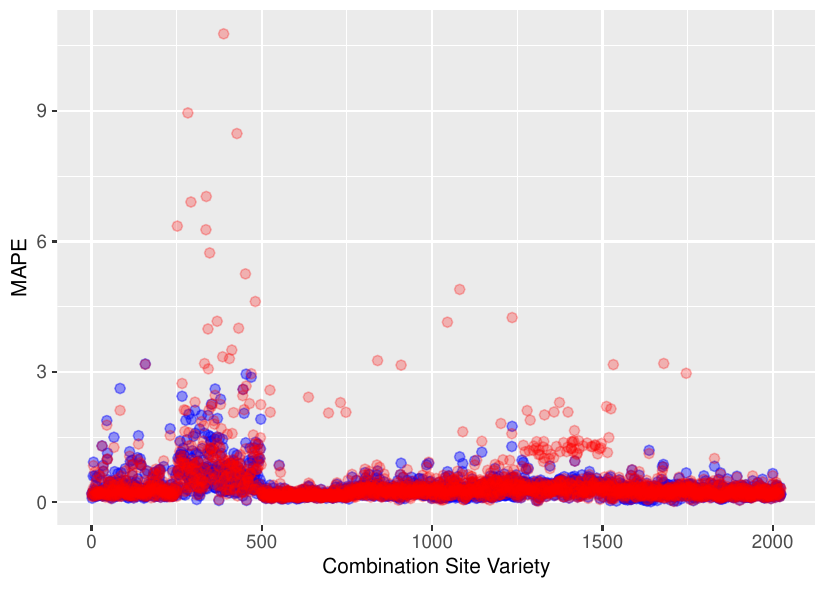}
  \caption{Imputation accuracy of the learned BNs, $\BLME$ (blue points) and
    $\BCGBN$ (red points), in terms of grain yield Mean Absolute Percentage
    Error (MAPE) of each site-variety combination, shown sequentially for
    brevity. Lower values are better.}
  \label{fig:imputation}
\end{figure}

The MAPE for the imputation of different site-variety combinations is shown in
Figure~\ref{fig:imputation}. We observe that $\BLME$ and $\BCGBN$ perform
similarly for all combinations except those involving the sites of Craiova
(numbered 250--500) and Campagnola (numbered 1250--1500), for which $\BCGBN$ has
a higher MAPE than $\BLME$.

\begin{figure}[H]
\centering
  \includegraphics[width=\textwidth]{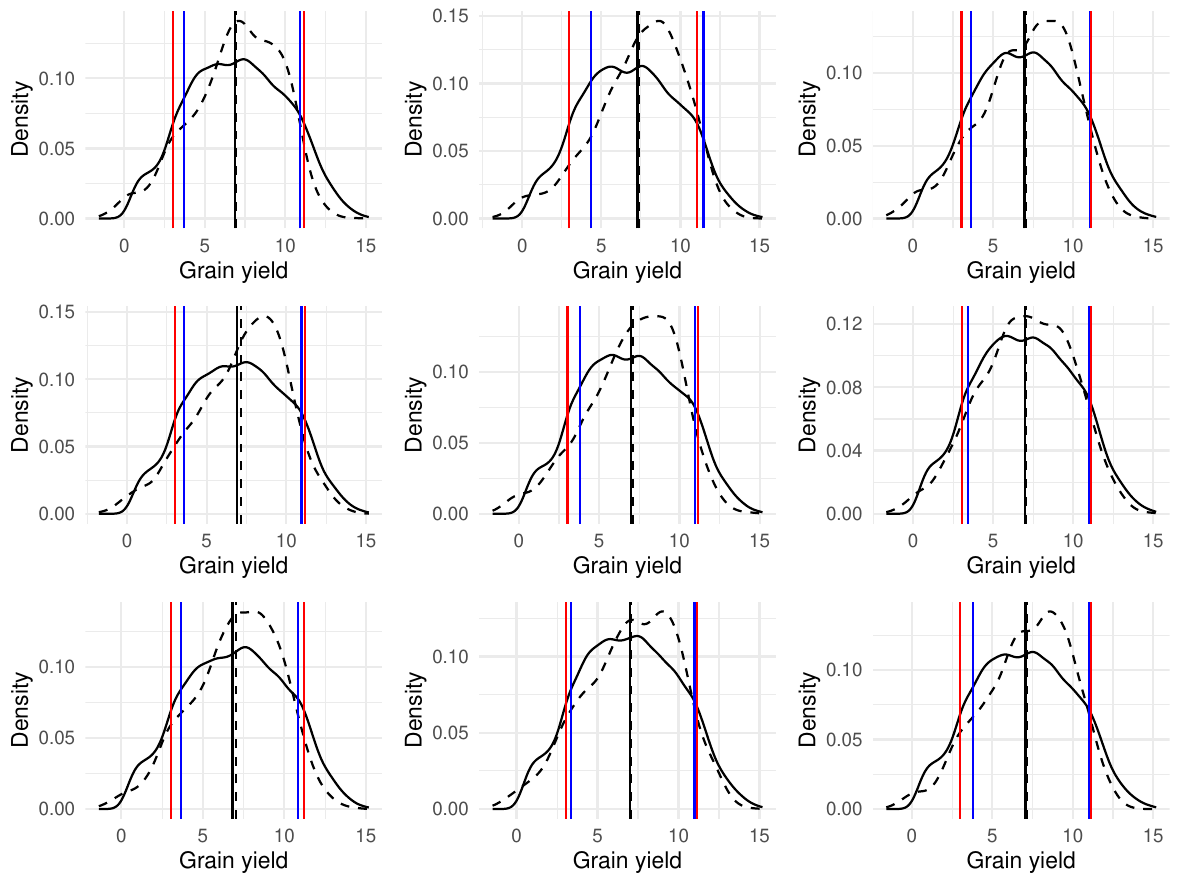}
  \caption{Kernel densities of the grain yield in the training set are
    represented by the solid curve, while the dashed curve depicts the kernel
    densities of the predicted grain yield obtained through likelihood-weighted
    approximation during cross-validation. The kernel density-based credible
    interval at 80\% for the grain yield in the training set is indicated by the
    red line and for the predicted grain yield by the blue line. The mean is
    reported with a solid line for the grain yield of the training set and a
    dashed line for the predicted grain yield.}
  \label{fig:kernel}
\end{figure}

The kernel densities of grain yield for the first nine runs of cross-validation
are shown in Figure~\ref{fig:kernel}, full results are reported as Supplementary
Materials. The densities for the training set exhibit somewhat heavier tails
than those for the predicted values. Furthermore, the predictive densities have
narrower credible intervals than those from the training set, particularly on
the lower tail, and are more often positively skewed. The mean values are nearly
identical for both, at approximately $7 t/ha$, with a $0.8$ credible interval
$[4t/ha, 11 t/ha]$.

Finally, we employed a stepwise parent elimination algorithm to search for
non-significant effects in each of the local distributions of the phenological
variables. The BIC values consistently indicated that, within $\BLME$, the best
set of effects were those selected by our method. The only exception was the
variable ``tassel height,'' for which the BIC was lower when the ``diurnal RH
range July--August'' variable was omitted. The same procedure was applied with
the removal of random effects from the local distributions. In this case, the
set of effects selected by our method still yielded the best BIC values.
Furthermore, we compared the BIC values for local distributions
with and without the random effects. Generally, the presence of random effects
improved goodness of fit, except for the variables ``tassel height'' and
``silking,'' which exhibited better BIC values in the absence of random effects.
All BIC results are reported in Table \ref{tab:mixed} and Table \ref{tab:fixed}.
The BIC values of the first row correspond to the set of parents of the local
distribution found with our method, and the other rows correspond to the BIC
value found after each parent elimination.

\section{Discussion}
\label{sec:discussion}

In this paper, we used a Bayesian network (BN) to analyse the results of a
multi-site agronomic experiment comprising eight different sites in Europe and
Chile. Our goal was to obtain a network model that can be used for causal
inference, thus providing an ideal foundation to develop decision support
systems to manage maize crops. To do that effectively, we modified the BN's
structure learning to encode the data's hierarchical structure, thus addressing
the violation of the exchangeability assumption that characterises the RDs.

The data we used consisted of weather variables and phenological variables of
maize measured from 2011 to 2013. In our study, we selected a subset of
variables based on their agronomic relevance in addition to the weather
variables for temperature and humidity. The latter was measured daily for each
site, so we calculated their mean for specific time-slices corresponding to the
key phenological phases of maize, namely seeding, germination, emergence
(May--June); vegetation stage, tasselling, silking, ear formation
(July--August); and grain filling, maturation and harvest (September--October).
The reason for this choice was to capture the effect of each weather variable on
the phenological variables. Based on strong prior knowledge, specific arcs were
prohibited due to their lack of causal meaning. For example, it is not plausible
for a weather variable from a later time slice to affect another in an earlier
time slice. We applied the same logical reasoning to the connections between
phenological variables recorded in different time slices. For instance, the arc
from grain yield to silking was prohibited, as it is causally impossible for the
grain yield to cause female flowering (silking). Moreover, all arcs that made
the cluster variable a child of other variables were prohibited.

We compared the structures of the BN incorporating random effects ($\BLME$) to
the baseline including only fixed effects ($\BCGBN$): the former contains 26
additional arcs compared to the latter, with a significant difference in the
case of grain yield, which had eight more parents. We further assessed the
predictive accuracy of phenological variables in both $\BLME$ and $\BCGBN$ using
the Diebold-Mariano test \citep{dmtest}. This statistical evaluation allowed us
to determine that the predictive accuracy improvement observed in $\BLME$ was
statistically significant for all of the variables (p-value $< 0.05$, results
not shown). We are aware that predictive accuracy is not an adequate measure of
performance for a causal model, which is better assessed using expert knowledge
as we do below, but it provides a term of comparison for our model in the wider
context of machine learning models, which cannot be assessed causally.

Regarding grain yield, plant height emerged as a new parent: its role as a
reliable predictor for maize grain yield is well-documented in the literature
\citep{yin_season_2011, pugh_temporal_2018}. Additionally, its ease of
measurement using remote sensing makes it a suitable candidate for predicting
maize grain yield \citep{remote2, remote_maize}. Supporting evidence comes from
the work of \citet{anderson_prediction_2019}, who studied 280 hybrids conducted
in 1,500 plots using unmanned aerial systems and found a positive correlation
between plant height and maize grain yield. Another new parent identified in the
analysis is silking. Existing evidence also supports this finding, as
\citet{malik_genetic_2005} demonstrated a significant negative correlation
between silking and grain yield. They posited that this negative relationship
could be attributed to late female flowering, resulting in a less favourable
photoperiod and low temperature induced by changing seasons. Considering
variables related to temperature and relative humidity (RH), they are all the
parents of grain yield in $\BLME$ but not in $\BCGBN$, where only diurnal RH
range May--June (RH4), diurnal RH range Sept--Oct (RH6), average temperature
May--June (T1), average temperature July--Aug (T2), diurnal temperature range
May--June (T4) where present. This is plausible since environmental conditions
are essential for maize growth: for instance, evidence shows that high
humidity during flowering promotes the maize yield
\citep{butts_wilmsmeyer_weather_2019}. Temperature plays a crucial role in
influencing maize yield, particularly during the reproductive phase, where
sub-optimal or supra-optimal values can have a significant impact. For instance,
temperatures ranging from 33°C to 36°C during the pre-and post-flowering stages
can result in a reduction of grain yield by 10\% to 45\%
\citep{neiff_high_2016}. In a review by \citet{waqas_thermal_2021}, the
detrimental effects of thermal stress on maize growth were thoroughly examined
from both an agronomic and a physiological perspective. They emphasised that
high temperatures, especially during the flowering period, can have various
adverse consequences on floret number, silk number, and grain development.
Furthermore, the process of fertilisation and grain-filling may also be
compromised under such conditions. On the other hand, low temperatures below
10°C can also be detrimental, negatively impacting the normal growth process of
maize. Such cold temperatures can limit germination, adversely affect root
morphology, and decrease the efficiency of photosystem II. These combined
factors demonstrate the sensitivity of maize to temperature fluctuations, which
can significantly influence its growth and overall productivity.

We applied hierarchical clustering to the mean and the variance of the residuals
from a simple linear regression of the grain yield, which was selected due to
its agronomic relevance against all other variables to avoid making any
assumptions about the possible parent grain yield. After grouping the residuals
by site-variety combination, hierarchical clustering produced 60
relatively-balanced clusters: they were included in the data as a discrete
variable that was set as a common parent of the phenological variables in a
setup similar to a conditional Gaussian BN as described in
Section~\ref{sec:learn}. We decided to use the clusters, rather than just the
site of origin or the maize variety as individual variables, for two reasons:
\begin{itemize}
  \item When using either the site or the maize variety as a common discrete
    parent variable, we found the dispersion of residuals in the local
    distribution, particularly that of grain yield, to be non-homogeneous.
  \item Combining the site of origin and the maize variety without clustering
    their combinations produces a variable with approximately 2000 possible
    values, which would make BN structure learning computationally prohibitive.
\end{itemize}
As a result, we improved the model's computational feasibility and predictive
accuracy. Using clustering as a pre-processing step has been proposed in the
literature to find suitable scenarios in risk assessment analysis
\citep{pettet_incident_2017} or to reduce the dimension of the estimation
problem, learning the structure of one subgraph for each cluster
\citep{gu_learning_2020}. \citet{multipartition_2022} also proposed a
multi-partition clustering that produces a set of categorical variables that
encode clusters. These partitions represent a distinct clustering solution and
were used as parents, leading to a more interpretable and flexible way to find
clusters.

As discussed in Section~\ref{sec:learn}, we assumed that the local distributions
of phenological variables are linear mixed-effect regression models to allow for
the partial pooling of information across clusters: this model balances the
individual cluster-specific estimates with the overall trend observed in the
data, leading to more stable and reliable estimates \citep{scutari2022}. We
assumed different local distributions for grain yield and the weather variables.
For grain yield, we introduced a power function to model the variance after
observing a skewed residual distribution during the exploratory data analysis.
Moreover, we modelled grain yield with a random intercept as the only random
effect; in contrast, all other phenological variables have both random
coefficients and intercepts. For the weather variables, we used a linear
regression model containing only fixed effects as the local distribution. We
made this decision based on visual inspection, which revealed that the weather
variables appeared disconnected from the clusters. This observation implies that
the values of these variables were independent of both the site of origin and
the variety of maize.

These assumptions reduced the prediction error for grain yield from 28\% to
approximately 17\% when its Markov blanket or its parents were used as
predictors, as shown in Figure~\ref{fig:accuracy}. In two specific
intervals, our model exhibits a higher Mean Absolute Percentage Error (MAPE)
compared to the baseline model. These intervals are scenarios 7 to 11 and
scenarios 19 to 24. In the first interval (7--11), our model's performance is
affected as we gradually introduce variables related to relative humidity (RH).
The second interval (19--24) corresponds to the gradual inclusion of phenotypic
variables. The reason for our model's higher MAPE in these ranges could indicate
that simply considering phenotypic and humidity variables is insufficient for
using our model as a reliable Decision Support System (DSS). However, a notable
trend emerges when we combine temperature and humidity variables (scenarios
13--18) and subsequently add phenotypic variables (24--32). During these
scenarios, the MAPE of our model significantly improves compared to the
baseline. Interestingly, this improvement does not occur with temperature alone,
where our model's MAPE remains lower than the baseline. This suggests that
introducing temperature as a new variable enhances the model's performance. This
observation aligns with the causal and biological context since temperature
plays a pivotal role in influencing the phenological stages of plants. Hence,
incorporating temperature as a factor results in more accurate predictions.
Furthermore, we assessed whether the incorporation of random effects in the
structure learning procedure enhanced the local distributions, not just in terms
of structure but also in terms of model specification using a backward algorithm
that removes variables from the local distribution and tests the BIC, results
are reported in Tables~\ref{tab:mixed} and~\ref{tab:fixed} in the Appendix.
\citet{millet_genomic_2019} used the same data set for grain yield prediction,
employing grain weight and grain number as predictors. Their approach included
modelling grain numbers through a factorial regression model with predictors
such as specific intercepted radiation, soil water potential, night temperature,
hybrids, and experimental location. Their cross-validation analysis involved
testing new hybrids already evaluated in previous experiments and vice versa.
The correlation between the observed and predicted grain yield ranged from 0.43
to 0.85 per experiment and 0.71 to 0.97 per hybrid. For new hybrids in tested
experiments, correlation results ranged from 0.21 to 0.71 per experiment and
0.66 to 0.96 per hybrid. Our study employed a different cross-validation scheme
than \citet{millet_genomic_2019} to exclude simultaneous sites and varieties
from the training set. This led to correlations between observed and predicted
values ranging from 0.86 to 0.90, with an average of 0.88. We designed our
sampling scheme based on clustering to group together sites and varieties with
similar grain yield characteristics. Consequently, our model was tested by
randomly removing site-variety combinations, simulating scenarios where an
agronomist queries the model for grain yield predictions of sites and varieties
akin to those used in model training for enhanced robustness assessment.

Our findings indicated that random effects have a favourable impact on both
structure learning and model specification. They contribute to a more accurate
explanation of the data without introducing undue complexity. However,
exceptions were observed for ``tassel height'' and ``silking'', which were best
modelled without random effects. This suggests that the maxima identified with
our method might be local maxima rather than global ones. Additionally, it
implies that the estimation of their local distributions did not benefit from
the partial pooling information provided by random effects.

Our findings confirm that a CGBN incorporating mixed-effects models to exploit
the hierarchical structure of the data provides better accuracy than a standard
CGBN. Considering that it is a causal model as well, we argue that it can serve
as an effective decision support system, particularly in domains with inherent
hierarchical structures, such as the agronomic field
\citep{burchfield_impact_2019,li_hierarchical_2020}.

Even though our proposed method exhibits low prediction error, it has
limitations. Firstly, the clustering pre-processing is based only on grain yield
regression to simplify the model and to reduce the computational cost of
learning it. To address this, future work will expand the clustering approach to
specific clusters for each phenological variable, enabling a more detailed
analysis. Another limitation is the time it took to learn it: approximately 13
hours of linear time. To mitigate this, we may explore different approaches to
model hierarchical data, such as the Integrated Nested Laplace Approximation
\citep[INLA; ][]{rue_bayesian_2017}. Moreover, in this study, we used the
hill-climbing algorithm for structure learning because it compares favourably in
terms of speed and structural accuracy \citep{scutari_who_2019} and because our
focus was on incorporating mixed effects. Hence, we wanted to avoid
hyper-parameter tuning issues common with more complex structure learning
algorithms. Other algorithms, however, may very well be more suitable for this
particular type of data, and we will explore them in future work. In particular,
reformulating structure learning to share information on the covariance
structure of the data between iterations and variables has the potential to
vastly reduce computational complexity without impacting accuracy.

\section{Conclusions and Future works}
\label{sec:conclusion}

Maize stands as a crucial crop for both food and feed production. Predicting its
yield can enhance farm practices and refine crop management systems. In
agricultural datasets, a common hierarchical structure can be leveraged for
predictive purposes; therefore, models that harness this structure tend to yield
highly accurate predictions. In this study, we introduced a novel approach to
learning Bayesian Networks integrating random effects into local distributions,
obtaining such conclusions:
\begin{itemize}
  \item The variables encoding the provenance of the field (\textit{Site})
    exhibit a dispersion of residuals in the local distribution.
  \item Introducing a group variable by performing hierarchical clustering based
    on the mean and variance of residuals from a complete linear regression of
    maize grain yield, conditioned on the site-variety combination, can reduce
    the dimension of the Cartesian product of site-Variety combinations.
  \item The application of this method led to a reduction in prediction errors
    compared to the baseline model.
  \item The reduction in errors was attributed to the partial pooling of
    information provided by the random effects.
\end{itemize}

We propose its applicability as a valuable decision support system, particularly
in fields marked by inherent hierarchical structures like agronomy. From an
agricultural engineering point of view, crop yield predictions provide good
management of agricultural practices, optimally scheduling the irrigation system
and phytosanitary treatments, leading farming systems resilient to climate
change and economic losses. Future work, as we explained in the discussion
section, will consider performing precise clustering for each phenotypical
variable. Additionally, introducing alternative estimation methods, such as
INLA, could reduce computational demands and empower experts to actively engage
in the learning process within the Bayesian framework. Moreover, such methods
could be used to take into account the spatial component of the datasets since
INLA is typically used for such purposes.

\section{Data availability}

The dataset is available online at the following link:
\url{https://entrepot.recherche.data.gouv.fr/dataset.xhtml?persistentId=doi:10.15454/IASSTN}.

\pagebreak

\appendix

\section{More details about Scenarios, Structure learning and Validation tests}
\label{sec:sample:appendix}

\begin{table}[ht]
  \caption{Prediction scenarios for the grain yield of maize, identified by the
    set of variables used as predictors: the average temperature May--June (T1),
    the average temperature July--Aug (T2), the average temperature Sept--Oct
    (T3), the diurnal temperature range May--June (T4), the diurnal temperature
    range July--Aug (T5), the diurnal temperature range Sept--Oct (T6), the
    average RH May--June (RH1), the average RH July--Aug (RH2), the average RH
    Sept--Oct (RH3), the diurnal RH range May--June (RH4), the diurnal RH range
    July--Aug (RH5), the diurnal RH range Sept--Oct (RH6), Silking (Si), GW
    (Grain weight), An (Anthesis), TH (Tassel height), PH (Plant height) and EH
    (Ear height).}
  \label{tab:scenarios}
  \resizebox{\columnwidth}{!}{%
  \begin{tabular}{ccccccccccccccccccc}
    \hline
    Scenario & T1 & T2 & T3 & T4 & T5 & T6 & RH1 & RH2 & RH3 & RH4 & RH5 & RH6 & Si & GW & TH & PH & An & EH \\ \hline
    1 & $\checkmark$ &  &  &  &  &  &  &  &  &  &  &  &  &  &  &  &  &  \\
    2 & $\checkmark$ & $\checkmark$ &  &  &  &  &  &  &  &  &  &  &  &  &  &  &  &  \\
    3 & $\checkmark$ & $\checkmark$ &  & $\checkmark$ &  &  &  &  &  &  &  &  &  &  &  &  &  &  \\
    4 & $\checkmark$ & $\checkmark$ &  & $\checkmark$ & $\checkmark$ &  &  &  &  &  &  &  &  &  &  &  &  &  \\
    5 & $\checkmark$ & $\checkmark$ &  & $\checkmark$ & $\checkmark$ & $\checkmark$ &  &  &  &  &  &  &  &  &  &  &  &  \\
    6 & $\checkmark$ & $\checkmark$ & $\checkmark$ & $\checkmark$ & $\checkmark$ & $\checkmark$ &  &  &  &  &  &  &  &  &  &  &  &  \\
    7 &  &  &  &  &  &  & $\checkmark$ &  &  &  &  &  &  &  &  &  &  &  \\
    8 &  &  &  &  &  &  & $\checkmark$ & $\checkmark$ &  &  &  &  &  &  &  &  &  &  \\
    9 &  &  &  &  &  &  & $\checkmark$ & $\checkmark$ &  & $\checkmark$ &  &  &  &  &  &  &  &  \\
    10 &  &  &  &  &  &  & $\checkmark$ & $\checkmark$ &  & $\checkmark$ & $\checkmark$ &  &  &  &  &  &  &  \\
    11 &  &  &  &  &  &  & $\checkmark$ & $\checkmark$ &  & $\checkmark$ & $\checkmark$ & $\checkmark$ &  &  &  &  &  &  \\
    12 &  &  &  &  &  &  & $\checkmark$ & $\checkmark$ & $\checkmark$ & $\checkmark$ & $\checkmark$ & $\checkmark$ &  &  &  &  &  &  \\
    13 & $\checkmark$ &  &  &  &  &  & $\checkmark$ &  &  &  &  &  &  &  &  &  &  &  \\
    14 & $\checkmark$ & $\checkmark$ &  &  &  &  & $\checkmark$ & $\checkmark$ &  &  &  &  &  &  &  &  &  &  \\
    15 & $\checkmark$ & $\checkmark$ &  & $\checkmark$ &  &  & $\checkmark$ & $\checkmark$ &  & $\checkmark$ &  &  &  &  &  &  &  &  \\
    16 & $\checkmark$ & $\checkmark$ &  & $\checkmark$ & $\checkmark$ &  & $\checkmark$ & $\checkmark$ &  & $\checkmark$ & $\checkmark$ &  &  &  &  &  &  &  \\
    17 & $\checkmark$ & $\checkmark$ &  & $\checkmark$ & $\checkmark$ & $\checkmark$ & $\checkmark$ & $\checkmark$ &  & $\checkmark$ & $\checkmark$ & $\checkmark$ &  &  &  &  &  &  \\
    18 & $\checkmark$ & $\checkmark$ & $\checkmark$ & $\checkmark$ & $\checkmark$ & $\checkmark$ & $\checkmark$ & $\checkmark$ & $\checkmark$ & $\checkmark$ & $\checkmark$ & $\checkmark$ &  &  &  &  &  &  \\
    19 &  &  &  &  &  &  &  &  &  &  &  &  & $\checkmark$ &  &  &  &  &  \\
    20 &  &  &  &  &  &  &  &  &  &  &  &  & $\checkmark$ & $\checkmark$ &  &  &  &  \\
    21 &  &  &  &  &  &  &  &  &  &  &  &  & $\checkmark$ & $\checkmark$ & $\checkmark$ &  &  &  \\
    22 &  &  &  &  &  &  &  &  &  &  &  &  & $\checkmark$ & $\checkmark$ & $\checkmark$ & $\checkmark$ &  &  \\
    23 &  &  &  &  &  &  &  &  &  &  &  &  & $\checkmark$ & $\checkmark$ & $\checkmark$ & $\checkmark$ & $\checkmark$ &  \\
    24 &  &  &  &  &  &  &  &  &  &  &  &  & $\checkmark$ & $\checkmark$ & $\checkmark$ & $\checkmark$ & $\checkmark$ & $\checkmark$ \\
    25 & $\checkmark$ &  &  &  &  &  & $\checkmark$ &  &  &  &  &  & $\checkmark$ &  &  &  &  &  \\
    26 & $\checkmark$ & $\checkmark$ &  &  &  &  & $\checkmark$ & $\checkmark$ &  &  &  &  & $\checkmark$ & $\checkmark$ &  &  &  &  \\
    27 & $\checkmark$ & $\checkmark$ &  & $\checkmark$ &  &  & $\checkmark$ & $\checkmark$ &  & $\checkmark$ &  &  & $\checkmark$ & $\checkmark$ & $\checkmark$ &  &  &  \\
    28 & $\checkmark$ & $\checkmark$ &  & $\checkmark$ & $\checkmark$ &  & $\checkmark$ & $\checkmark$ &  & $\checkmark$ & $\checkmark$ &  & $\checkmark$ & $\checkmark$ & $\checkmark$ & $\checkmark$ &  &  \\
    29 & $\checkmark$ & $\checkmark$ &  & $\checkmark$ & $\checkmark$ & $\checkmark$ & $\checkmark$ & $\checkmark$ &  & $\checkmark$ & $\checkmark$ & $\checkmark$ & $\checkmark$ & $\checkmark$ & $\checkmark$ & $\checkmark$ & $\checkmark$ &  \\
    30 & $\checkmark$ & $\checkmark$ & $\checkmark$ & $\checkmark$ & $\checkmark$ & $\checkmark$ & $\checkmark$ & $\checkmark$ & $\checkmark$ & $\checkmark$ & $\checkmark$ & $\checkmark$ & $\checkmark$ & $\checkmark$ & $\checkmark$ & $\checkmark$ & $\checkmark$ & $\checkmark$ \\
    31 & $\checkmark$ & $\checkmark$ &  & $\checkmark$ & $\checkmark$ & $\checkmark$ & $\checkmark$ & $\checkmark$ & $\checkmark$ & $\checkmark$ & $\checkmark$ & $\checkmark$ & $\checkmark$ & $\checkmark$ & $\checkmark$ &  &  &  \\
    32 & $\checkmark$ & $\checkmark$ &  & $\checkmark$ & $\checkmark$ & $\checkmark$ & $\checkmark$ & $\checkmark$ & $\checkmark$ & $\checkmark$ &  &  & $\checkmark$ &  & $\checkmark$ &  &  &  \\ \hline
  \end{tabular}%
  }
\end{table}

\begin{table}[p]
  \centering
  \caption{New relationships found in $\BLME$. Variables: the average
    temperature May--June (T1), the average temperature July--Aug (T2), the
    average temperature Sept--Oct (T3), the diurnal temperature range May--June
    (T4), the diurnal temperature range July--Aug (T5), the diurnal temperature
    range Sept--Oct (T6), the average RH May--June (RH1), the average RH
    July--Aug (RH2), the average RH Sept--Oct (RH3), the diurnal RH range
    May--June (RH4), the diurnal RH range July--Aug (RH5), the diurnal RH range
    Sept--Oct (RH6), Silking (Si), GW (Grain weight), An (Anthesis), TH (Tassel
    height), PH (Plant height) and EH (Ear height).}
  \label{tab:arc}
  \begin{tabular}{lll}
    \hline
    Parent &  & Child \\
    \hline
    PH & $\rightarrow$ & GY \\
    PH & $\rightarrow$ & EH \\
    EH & $\rightarrow$ & Si \\
    Si & $\rightarrow$ & GY \\
    T1 & $\rightarrow$ & EH \\
    T1 & $\rightarrow$ & PH \\
    T2 & $\rightarrow$ & An \\
    T2 & $\rightarrow$ & TH \\
    T3 & $\rightarrow$ & GY \\
    T4 & $\rightarrow$ & GW \\
    T4 & $\rightarrow$ & Si \\
    T4 & $\rightarrow$ & TH \\
    T5 & $\rightarrow$ & GY \\
    \hline
  \end{tabular}
  \hspace{4\baselineskip}
  \begin{tabular}{lll}
    \hline
    Parent &  & Child \\ \hline
    T5 & $\rightarrow$ & Si \\
    T5 & $\rightarrow$ & TH \\
    T5 & $\rightarrow$ & PH \\
    T6 & $\rightarrow$ & GW \\
    RH1 & $\rightarrow$ & GY \\
    RH2 & $\rightarrow$ & GY \\
    RH3 & $\rightarrow$ & GY \\
    RH4 & $\rightarrow$ & TH \\
    RH4 & $\rightarrow$ & PH \\
    RH5 & $\rightarrow$ & GY \\
    RH5 & $\rightarrow$ & EH \\
    RH5 & $\rightarrow$ & PH \\
    RH6 & $\rightarrow$ & GW \\
    \hline
  \end{tabular}
\end{table}

\begin{table}[p]
  \caption{The BIC score values for the variables with random effects in
  $\BLME$. The columns correspond to Silking (Si), GW (Grain weight), An
  (Anthesis), TH (Tassel height), PH (Plant height), EH (Ear height) and GY
  (Grain yield).}
  \label{tab:mixed}
  \begin{tabular}{cllllll}
    \hline
    \multicolumn{1}{l}{EH} & PH & TH & An & Si & GW & GY \\ \hline
    \multicolumn{1}{l}{108448.0} & 128945.4 & 98210.88 & 72373.64 & 64279.80 & 127634.6 & 42664.90 \\
    \multicolumn{1}{l}{108581.7} & 128988.4 & 131629.19 & 73127.37 & 67958.98 & 129620.1 & 42788.77 \\
    \multicolumn{1}{l}{108913.9} & 130979.4 & 98721.24 & 72466.75 & 64423.46 & 130956.2 & 43600.06 \\
    \multicolumn{1}{l}{109422.9} & 129204.0 & 98306.37 & 73802.03 & 78577.20 & 129137.0 & 42817.61 \\
    \multicolumn{1}{l}{109076.0} & 128987.3 & 98975.48 & 73569.12 & 64617.86 & 128784.5 & 47312.45 \\
    \multicolumn{1}{l}{108793.3} & 128996.4 & 99125.48 & 74726.54 & 64740.27 & 129580.9 & 42683.18 \\
    - & 130634.3 & 98957.10 & 73009.51 & 65812.12 & 127641.1 & 44288.90 \\
    - & \multicolumn{1}{c}{-} & 98134.49 & 74775.35 & 65272.43 & 129603.5 & 43055.93 \\
    - & \multicolumn{1}{c}{-} & \multicolumn{1}{c}{-} & 75937.24 & 64331.62 & 130433.6 & 42733.74 \\
    - & \multicolumn{1}{c}{-} & \multicolumn{1}{c}{-} & \multicolumn{1}{c}{-} & 64426.97 & 128022.4 & 43319.44 \\
    - & \multicolumn{1}{c}{-} & \multicolumn{1}{c}{-} & \multicolumn{1}{c}{\textbf{-}} & \multicolumn{1}{c}{-} & 129434.5 & 43919.43 \\
    - & \multicolumn{1}{c}{-} & \multicolumn{1}{c}{-} & \multicolumn{1}{c}{-} & \multicolumn{1}{c}{-} & \multicolumn{1}{c}{-} & 43890.76 \\
    - & \multicolumn{1}{c}{-} & \multicolumn{1}{c}{-} & \multicolumn{1}{c}{-} & \multicolumn{1}{c}{-} & \multicolumn{1}{c}{-} & 44146.02 \\
    - & \multicolumn{1}{c}{-} & \multicolumn{1}{c}{-} & \multicolumn{1}{c}{-} & \multicolumn{1}{c}{-} & \multicolumn{1}{c}{-} & 42899.58 \\
    - & \multicolumn{1}{c}{-} & \multicolumn{1}{c}{-} & \multicolumn{1}{c}{-} & \multicolumn{1}{c}{-} & \multicolumn{1}{c}{-} & 42926.77 \\ \hline
  \end{tabular}
\end{table}

\begin{table}[p]
  \caption{The BIC score values for the variables without random effects in
    $\BLME$. The columns correspond to Silking (Si), GW (Grain weight), An
    (Anthesis), TH (Tassel height), PH (Plant height), EH (Ear height) and GY
    (Grain yield).}

  \label{tab:fixed}
  \begin{tabular}{cllllll}
  \hline
  \multicolumn{1}{l}{EH} & PH & TH & An & Si & GW & GY \\ \hline
  \multicolumn{1}{l}{109382.7} & 129069.7 & 97997.56 & 72921.13 & 63953.28 & 128026.4 & 50158.96 \\
  \multicolumn{1}{l}{109517.0} & 129130.9 & 132132.71 & 73968.06 & 67636.88 & 129661.0 & 50252.83 \\
  \multicolumn{1}{l}{110087.1} & 131885.7 & 99046.89 & 73012.02 & 64104.01 & 131334.4 & 50586.06 \\
  \multicolumn{1}{l}{109464.6} & 129540.4 & 98415.05 & 74703.04 & 78859.63 & 129263.1 & 50196.68 \\
  \multicolumn{1}{l}{109928.2} & 129201.7 & 99377.01 & 74227.08 & 64271.16 & 130586.0 & 53345.75 \\
  \multicolumn{1}{l}{109577.8} & 129234.8 & 99329.96 & 75441.82 & 64386.79 & 131320.9 & 50161.03 \\
  - & 131600.8 & 99481.83 & 73857.10 & 65460.04 & 128286.2 & 51321.11 \\
  - & \multicolumn{1}{c}{-} & 98304.39 & 75487.06 & 65002.78 & 130356.2 & 50378.48 \\
  - & \multicolumn{1}{c}{-} & \multicolumn{1}{c}{-} & 76760.46 & 64166.55 & 132314.7 & 50178.26 \\
  - & \multicolumn{1}{c}{-} & \multicolumn{1}{c}{-} & \multicolumn{1}{c}{-} & 64281.80 & 128579.4 & 50610.04 \\
  - & \multicolumn{1}{c}{-} & \multicolumn{1}{c}{-} & \multicolumn{1}{c}{-} & \multicolumn{1}{c}{-} & 130592.4 & 51098.38 \\
  - & \multicolumn{1}{c}{-} & \multicolumn{1}{c}{-} & \multicolumn{1}{c}{-} & \multicolumn{1}{c}{-} & \multicolumn{1}{c}{-} & 51108.30 \\
  - & \multicolumn{1}{c}{-} & \multicolumn{1}{c}{-} & \multicolumn{1}{c}{-} & \multicolumn{1}{c}{-} & \multicolumn{1}{c}{-} & 50974.64 \\
  - & \multicolumn{1}{c}{-} & \multicolumn{1}{c}{-} & \multicolumn{1}{c}{-} & \multicolumn{1}{c}{-} & \multicolumn{1}{c}{-} & 50306.71 \\
  - & \multicolumn{1}{c}{-} & \multicolumn{1}{c}{-} & \multicolumn{1}{c}{-} & \multicolumn{1}{c}{-} & \multicolumn{1}{c}{-} & 50364.45 \\ \hline
  \end{tabular}
\end{table}

\pagebreak

% \bibliographystyle{elsarticle-harv}
% \bibliography{references}

\end{document}